\documentclass[11pt,a4paper]{article}
\usepackage[hyperref]{naaclhlt2018}
\usepackage{times}
\usepackage{latexsym}
\usepackage{arabtex}
\usepackage{lipsum}
\usepackage{url}
\usepackage{booktabs}
\usepackage{graphicx}
\usepackage{multirow}
\usepackage{arydshln}

\aclfinalcopy 

\newcommand{\Ni}{({\em i})~}
\newcommand{\Nii}{({\em ii})~}
\newcommand{\Niii}{({\em iii})~}
\newcommand{\Niv}{({\em iv})~}

\title{Integrating Stance Detection and Fact Checking in a Unified Corpus}

\author{
	Ramy Baly$^1$, Mitra Mohtarami$^1$, James Glass$^1$ \\
    {\bf Llu\'is M\`arquez$^{3}$\hspace{-.3em}~\thanks{ \hspace{.3em} This work was carried out when the authors were scientists at QCRI, HBKU.}~~, Alessandro Moschitti$^{3\ast}$, Preslav Nakov$^2$}\\
    $^1$MIT Computer Science and Artificial Intelligence Laboratory, MA, USA\\
    $^2$Qatar Computing Research Institute, HBKU, Qatar; $^3$Amazon\\
    {\tt \{baly,mitram,glass\}@mit.edu}\\
    {\tt \{lluismv,amosch\}@amazon.com; pnakov@qf.org.qa}
 }

\date{}

\begin{document}
\maketitle

\centerline{\large\bf Abstract}
\begin{table}[ht]
\centering
\scalebox{0.92}{
\begin{tabular}{p{2.8in}}
A reasonable approach for fact checking a claim involves
retrieving potentially relevant documents from different sources (e.g.,~news websites, social media, etc.),
determining the stance of each document with respect to the claim, and finally
making a prediction about the claim's factuality by aggregating the strength of the stances, while taking the reliability of the source into account.
Moreover, a fact checking system should be able to explain its decision by providing relevant extracts (rationales) from the documents.
Yet, this setup is not directly supported by existing datasets, which treat fact checking, document retrieval, source credibility, stance detection and rationale extraction as independent tasks.
In this paper, we support the interdependencies between these tasks as annotations in the same corpus.
We implement this setup on an Arabic fact checking corpus, the first of its kind.
\end{tabular}}
\end{table}

\section{Introduction}

Fact checking has recently emerged as an important research topic due to the unprecedented amount of fake news and rumors that are flooding the Internet in order to manipulate people's opinions \cite{SeminarUsers2017,Mihaylov2015FindingOM,Mihaylov2015ExposingPO,mihaylov-nakov:2016:P16-2} or to influence the outcome of major events such as political elections~\cite{Lazer1094,Vosoughi1146}.
While the number of organizations performing fact checking is growing,
these efforts cannot keep up with the pace at which false claims are being produced, including also  clickbait~\cite{RANLP2017:clickbait}, hoaxes \cite{rashkin2017truth}, and satire \cite{Hardalov2016}.
Hence, there is need for automation.

\noindent While most previous research has focused on English, here we target Arabic. Moreover, we propose some guidelines, which we believe should be taken into account when designing fact-checking corpora, irrespective of the target language.

Automatic fact checking typically involves retrieving potentially relevant documents (news articles, tweets, etc.), determining the stance of each document with respect to the claim, and finally predicting the claim's factuality by aggregating the strength of the different stances, taking into consideration the reliability of the documents' sources (news medium, Twitter account, etc.).
Despite the interdependency between \emph{fact checking} and \emph{stance detection}, research on these two problems has not been previously supported by an integrated corpus. This is a gap we aim to bridge by retrieving documents for each claim and annotating them for stance, thus ensuring a natural distribution of the stance labels.

Moreover, in order to be trusted by users, a fact-checking system should be able to explain the reasoning that led to its decisions.
This is best supported by showing extracts (such as sentences or phrases) from the retrieved documents that illustrate the detected stance \cite{lei-barzilay-jaakkola:2016:EMNLP2016}. Unfortunately, existing datasets do not offer manual annotation of sentence- or phrase-level supporting evidence.
While deep neural networks with attention mechanisms can infer and extract such evidence automatically in an unsupervised way \cite{parikh-EtAl:2016:EMNLP2016}, potentially better results can be achieved when having the target sentence provided in advance, which enables supervised or semi-supervised training of the attention.
This would allow not only more reliable evidence extraction, but also better stance prediction, and ultimately better factuality prediction.
Following this idea, our corpus also identifies the most relevant stance-marking sentences.

\section{Related Work}\label{sec:related_work}

The connection between fact checking and stance has been argued for by~\newcite{vlachos2014fact}, who envisioned a system that
\Ni~identifies factual statements~\cite{hassan2015detecting,gencheva2017context,NAACL2018:claimrank},
\Nii~generates questions or queries \cite{karadzhov2017fully},
\Niii~creates a knowledge base using information extraction and question answering~\cite{Ba:2016:VERA,shiralkar2017finding}, and
\Niv~infers the statements' veracity using text analysis~\cite{banerjee-han:2009:NAACLHLT09-Short,Castillo:2011:ICT:1963405.1963500,rashkin2017truth}
or information from external sources~\cite{popat2016credibility,karadzhov2017fully,Popat:2017:TLE:3041021.3055133}.
This connection has been also used in practice, e.g.,~by \newcite{Popat:2017:TLE:3041021.3055133}; however, different datasets had to be used for stance detection vs. fact checking, as no dataset so far has targeted both.

{\it Fact checking} is very time-consuming, and thus most datasets focus on claims that have been already checked by experts on specialized sites such as
{\it Snopes}~\cite{ma2016detecting,popat2016credibility,Popat:2017:TLE:3041021.3055133}, {\it PolitiFact}~\cite{wang:2017:Short}, or
{\it Wikipedia hoaxes}~\cite{popat2016credibility}.\footnote{Annotating from scratch is needed in some cases, e.g.,~in the context of question answering~\cite{AAAIFactChecking2018}, or when targeting credibility \cite{Castillo:2011:ICT:1963405.1963500}.}
As fact checking is mainly done for English, non-English datasets are rare and often unnatural, e.g.,~translated from English, and  focusing on US politics.\footnote{See for example the CLEF-2018 lab on Automatic Identification and Verification of Claims in Political Debates, which features US political debates translated to Arabic:\\\url{http://alt.qcri.org/clef2018-factcheck/}}
In contrast, we start with claims that are not only relevant to the Arab world, but that were also originally made in Arabic, thus producing the first publicly available Arabic fact-checking dataset.

{\it Stance detection} has been studied so far disjointly from fact checking.
While there exist some datasets for Arabic~\cite{DarwishMZ17}, the most popular ones are for English, e.g., from SemEval-2016 Task~6~\cite{mohammad2016semeval} and from the Fake News Challenge (FNC).\footnote{\url{http://www.fakenewschallenge.org/}}
Despite its name, 
the latter has no annotations for factuality, but 
consists of article-claim pairs labeled for stance: {\it agrees}, {\it disagrees}, {\it discusses}, and {\it unrelated}.
In contrast, we retrieve documents for each claim, which yields an arguably more natural distribution of stance labels compared to FNC.

\emph{Evidence extraction}. Finally, an important characteristic of our dataset is that it provides evidence, in terms of text fragments, for the \emph{agree} and \emph{disagree} labels. Having such supporting evidence annotated enables both better learning for supervised systems performing stance detection or fact checking, and also the ability for such systems to learn to explain their decisions to users.
Having this latter ability has been recognized in previous work on rationalizing neural predictions \cite{lei-barzilay-jaakkola:2016:EMNLP2016}. This is also at the core of recent research on machine comprehension, e.g.,~using the SQuAD dataset~\cite{rajpurkar-EtAl:2016:EMNLP2016}.
However, such annotations have not been done for stance detection or fact checking before.

Finally, while preparing the camera-ready version of the present paper, we came to know about a new dataset for Fact Extraction and VERification, or FEVER ~\cite{thorne2018fever},
which is somewhat similar to ours as it it about both factuality and stance, and it has annotation for evidence. Yet, it is also different as \Ni the claims are artificially generated by manually altering Wikipedia text, \Nii the knowledge base is restricted to Wikipedia articles, and \Niii the stance and the factuality labels are identical, assuming that Wikipedia articles are reliable to be able to decide a claim's veracity.
In contrast, we use real claims from news outlets, we retrieve articles from the entire Web, and we keep stance and factuality as separate labels.

\section{The Corpus}
\label{sec:corpus_development}

Our corpus contains claims labeled for factuality ({\it true} vs. {\it false}).
We associate each claim with several documents, where each claim-document pair is labeled for stance ({\it agree}, {\it disagree}, {\it discuss}, or {\it unrelated}) similar to the FakeNewsChallenge (FNC) dataset.
Overall, the process of corpus creation went through several stages -- {\it claim extraction}, {\it evidence extraction} and {\it stance annotation} --, which we describe below.

\paragraph{Claim Extraction}
We consider two websites as the source of our claims.
{\sc Verify}\footnote{\url{http://www.verify-sy.com}} is a project that was established to expose false claims made about the war in Syria and other related Middle Eastern issues.
It is an independent platform that debunks claims made by all parties to the conflict.
To the best of our knowledge, this is the only platform that publishes fact-checked claims in Arabic.

\noindent It is worth noting that the {\sc Verify} website only shows claims that were debunked as false and misleading, and hence we used it to extract only the \emph{false} claims for our corpus (we extracted the \emph{true} claims from a different source; see below).

We thoroughly preprocessed the original claims. First, we manually identified and excluded all claims discussing falsified multimedia (images or video), which cannot be verified using textual information and NLP techniques only, e.g.

\begin{quote}
(1) {\it Pro-regime pages have circulated pictures of fighters fleeing an explosion.}\\
\RL{n^srt .sf.hAt mwAlyT liln.zAm .swr lmqAtlyn yhrbwn mn AnfjAr}
\end{quote}

Note that the claims in {\sc Verify} were written in a form that presents the {\it corrected} information after debunking the original false claim.
For instance, the original false claim in example~2a is corrected and published in {\sc Verify} as shown in example~2b.
We manually rendered these corrected claims to their original false form, which we used for our corpus.

\begin{quote}
(2a) (original false claim) {\it FIFA intends to investigate the game between Syria and Australia.}\\
\RL{AlfyfA t`tzm Alt.hqyq fy AlmbArAT byn swryA wAstrAlyA}\\\\
(2b) (corrected claim in {\sc Verify}) {\it FIFA does not intend to investigate the game between Syria and Australia, as pro-regime pages claim.}\\
\RL{AlfyfA lA t`tzm Alt.hqyq fy AlmbArAT byn swryA wAstralyA, kmA td`y .sf.hAt mwAlyT liln.zAm.}
\end{quote}

After extracting the false claims from {\sc Verify}, we collected the true claims of our corpus from {\sc Reuters}\footnote{\url{http://ara.reuters.com}} by extracting headlines of news documents.
We used a list of manually selected keywords to extract claims with the same topics as those extracted from {\sc Verify}.

\noindent Then, we manually excluded claims that contained political rhetorical statements (see example~3 below), multiple facts, accusations or denials, and ultimately we only kept those claims that discuss factual events, i.e., that can be verified.

\begin{quote}
(3) {\it Presidents Vladimir Putin and Recep Tayyip Erdogan hope that Astana talks will lead to peace.}\\
\RL{Alr'iysAn flAdymyr bwtyn wrjb .tyyb Ardw.gAn ya'mlAn ba'n m.hAd_tAt AstAnT swf tu'dy AlY AlslAm.}
\end{quote}

Overall, starting with 1,381 claims, we ended up with 422 worth-checking claims:
219 {\it false} claims from {\sc Verify}, and 203 {\it true} claims from {\sc Reuters}.

\paragraph{Evidence Extraction} 
Following the assumption that identifying stance towards claims can help predict their veracity, we want to associate each claim with supporting and opposing pieces of textual evidence.
We used the Google custom search API
for document retrieval, and we performed the following steps to increase the likelihood of retrieving relevant documents.
First, as in~\cite{karadzhov2017fully}, we transformed each claim into sub-queries by selecting named entities, adjectives, nouns and verbs with the highest TF.DF score, calculated on a collection of documents from the claims' sources.
Then, we used these sub-queries with the claim itself as input to the search API and retrieved the first 20 returned links, from which we excluded those directing to {\sc Verify} and {\sc Reuters}, and social media websites that are mostly opinionated.
Finally, we calculated two similarity measures between the links' content (documents) and the claims: the tri-gram containment~\cite{lyon2001detecting} and the cosine distance between average word embeddings of both texts.\footnote{Word embeddings were generated by training the GloVe~\cite{pennington2014glove} model on the Arabic Gigaword~\cite{parker2011gigaword}}.
We only kept documents with non-zero values for both measures, yielding 3,042 documents: 1,239 for false claims and 1,803 for true claims.

\paragraph{Stance Annotation:} We used CrowdFlower to recruit Arabic speakers to annotate the claim-document pairs for stance.
Each pair was assigned to 3--5 annotators, who were asked to assign one of the following standard labels (also used at FNC): {\it agree}, {\it disagree}, {\it discuss} and {\it unrelated}.
First, we conducted small-scale pilot tasks to fine-tune the guidelines and to ensure their clarity.
The annotators were also asked to focus on the stance of the document towards the claim, regardless of the factuality of either text. This ensures that stance is captured without bias, so it can be used later with other information (e.g., time, website's credibility, author reliability) to predict factuality.
Finally, the annotators were asked to specify segments in the documents representing the {\it rationales} that made them assign {\it agree} or {\it disagree} as labels.
For quality control purposes, we further created a small hidden test set by annotating 50 pairs ourselves, and we used it to monitor the annotators' performance, keeping only those who maintained an accuracy of over 75\%.

Ultimately, we used majority voting to aggregate stance labels for each pair, using the annotators' performance scores to break ties.
On average, 77\% of the annotators for each claim-document pair agreed on its label, thus allowing proper majority aggregation for most pairs.
A total of 133 pairs with significant annotation disagreement required us to manually check and correct the proposed annotations.
We further automatically refined the documents by
\Ni~excluding sentences with more than 200 words, and \Nii~limiting the size of a document to 100 sentences. Such extra-long documents tend to originate from crawling ill-structured websites, or from parsing some specific types of websites such as web forums.

Table~\ref{tab:distribution} shows the distribution over the stance labels,\footnote{The corpus is available at \url{http://groups.csail.mit.edu/sls/downloads/}\\ and also at \url{http://alt.qcri.org/resources/}} which turns out to be very similar to that for the FNC dataset.
We can see that there are very few documents disagreeing with \emph{true} claims (about 0.5\%), which suggests that stance is positively correlated with factuality.
However, the number of documents agreeing with \emph{false} documents is larger than the number of documents disagreeing with them, which illustrates one of the main challenges when trying to predict the factuality of news based on stance.

\begin{table}[t]
\centering
\scalebox{0.74}{
\begin{tabular}{@{}l|c|c@{ }@{ }c@{ }@{ }c@{ }@{ }c@{}}
\toprule
\multirow{2}*{\bf Claims} & \multirow{2}*{\begin{tabular}{c}{\bf Annotated}\\{\bf Documents}\end{tabular}}
 & \multicolumn{4}{c}{\bf Stance (document-to-claim)} \\
 & & {\it Agree} & {\it Disagree} & {\it Discuss} & {\it Unrelated} \\ \midrule
False: 219 & 1,239 & 103 & 82 & 159 & 895 \\
True: 203 & 1,803 & 371 & 5 & 250 & 1,177 \\ \midrule
Total: 402 & 3,042 & 474 & 87 & 409 & 2,072 \\
\bottomrule
\end{tabular}}
\caption{Statistics about stance and factuality labels.\label{tab:distribution}}
\end{table}

\section{Experiments and Evaluation}\label{sec:results}

We experimented with our Arabic corpus, after preprocessing it with ATB-style segmentation using MADAMIRA~\cite{pasha2014madamira}, 
using the following systems:

\begin{itemize}
	\item {\sc FNC baseline system}. This is the FNC organizers' system, which trains a gradient boosting classifier using hand-crafted features reflecting polarity, refute, similarity and overlap between the document and the claim.

\item {\sc Athene.} It was second at FNC \cite{hanselowski2017athene}, and was based on a multi-layer perceptron with the baseline system's features, word $n$-grams, and features generated using latent semantic analysis and other factorization techniques.

\item {\sc UCL.} It was third at FNC~\cite{riedel2017simple}, training a softmax layer using similarity features.

\item {\sc Memory Network.} We also experimented with an end-to-end memory network that showed state-of-the-art results on the FNC data \cite{mitra2018memory}.

\end{itemize}

The evaluation results are shown in Table~\ref{tab:results}. 
We use 5-fold cross-validation, where all claim-document pairs for the same claim are assigned to the same fold.
We report \emph{accuracy}, macro-average \emph{$F_1$-score}, and \emph{weighted accuracy}, which is the official evaluation metric of FNC.

\begin{table*}[tbh]
\centering
\scalebox{0.78}{
\begin{tabular}{l|l|ccc:c}
\toprule
\bf Model & \bf document Content Used & \bf Weigh. Acc. & \bf Acc. & \bf F$_1$ (macro) & \bf F$_1$ ({\it agree, disagree, discuss, unrelated})\\ \midrule
Majority class & \multicolumn{1}{|c|}{---} & 34.8 & 68.1 & 20.3 & 0 / 0 / 0 / 81 \\ \midrule
\multirow{4}*{FNC baseline system} & full document (default) & 55.6 & 72.4 & 41.0 & 60.4 / 9.0 / 10.4 / 84 \\
 & best sentence & 50.5 & 70.6 & 37.2 & 50.3 / 5.4 / 10.3 / 82.9\\
 & best sentence {\bf +rationale} & 60.6 & 75.6 & 45.9 & 73.5 / 13.2 / 11.3 / 85.5\\
 & full document {\bf +rationale} & 64.8 & 78.4 & 53.2 & 84.4 / 32.5 / 8.4 / 87.5 \\ \midrule
\multirow{4}*{UCL {\it \small (\#3rd in FNC)}} & full document (default) & 49.3 & 66.0 & 37.1 & 47.0 / 7.8 / 13.4 / 80 \\
 & best sentence & 46.8 & 66.7 & 34.7 & 44.3 / 3.5 / 11.4 / 79.8 \\
 & best sentence {\bf +rationale} & 58.5 & 71.9 & 44.8 & 71.6 / 12.6 / 12.4 / 82.6 \\
 & full document {\bf +rationale} & 63.7 & 76.3 & 51.6 & 84.2 / 21.4 / 15.3 / 85.3 \\ \midrule
\multirow{4}*{Athene {\it\small (\#2rd in FNC)}} & full document (default) & 55.1 & 70.5 & 41.3 & 59.1 / 9.2 / 14.1 / 82.3 \\
 & best sentence & 48.0 & 67.5 & 36.1 & 43.9 / 4.00 / 15.7 / 80.7 \\
 & best sentence {\bf +rationale} & 60.6 & 74.3 & 48.0 & 73.5 / 18.2 / 15.9 / 84.6 \\
 & full document {\bf +rationale} & 65.5 & 80.2 & 55.8 & 85.0 / 36.6 / 12.8 / 88.8 \\
\midrule
\multirow{4}*{Memory Network} & full document (default) & 55.3 & 70.9 & 41.6 & 60.0 / 15.0 / 8.5 / 83.1 \\
 & best sentence & 52.4 & 71.0 & 38.2 & 58.1 / 8.1 / 4.1 / 82.6 \\
 & best sentence {\bf +rationale} & 60.1 & 75.5 & 46.4 & 72.5 / 23.1 / 4.1 / 85.7 \\
 & full document {\bf +rationale} & 65.8 & 79.7 & 55.2 & 86.9 / 31.3 / 14.9 / 87.6 \\
\bottomrule
\end{tabular}}
\caption{Performance of some stance detection models from FNC when applied to our Arabic corpus.}\label{tab:results}
\end{table*}

Overall, our corpus appears to be much harder than FNC. 
For instance, the FNC baseline system achieves weighted accuracy of 75.2 on FNC vs. 55.6 (up to 64.8) on our corpus.
We believe that this is because we used a realistic information retrieval approach (see Section~\ref{sec:corpus_development}),
whereas the FNC corpus contains a significant number of totally unrelated document--claim pairs, e.g.,~about 40\% of the {\it unrelated} examples have no word overlap with the claim (even after stemming!), which makes it much easier to correctly predict the {\it unrelated} class (and this class is also by far the largest).

\noindent Table~\ref{tab:results} allows us to study the utility of having gold rationales for the stance (for the \emph{agree} and \emph{disagree} classes only) under different scenarios.
First, we show the results when using the full document along with the claim, which is the default representation.
Then, we use the best sentence from the document, i.e., the one that is most similar to the claim as measured by the cosine of their average word embeddings. This performs worse, which can be attributed to sometimes selecting the wrong sentence.
%
Next, we experiment with using the rationale instead of the best sentence when applicable (i.e.,~for \emph{agree} and \emph{disagree}), while still using the best sentence for \emph{discuss} and \emph{unrelated}. 
This yields sizable improvements on all evaluation metrics, compared to using the best sentence (5-12 point absolute) or the full document (3-9 points absolute).
We further evaluate the impact of using the rationales, when applicable, but using the full document otherwise.
This setting performed best (80.2\% accuracy with {\sc Athene}, and 3-8 points of improvement over best+rationale), as it has access to most information: full document + rationale.

Overall, the above experiments demonstrate that having a gold rationale can enable better learning. However, the results should be considered as a kind of upper bound on the expected performance improvement, since here we used gold rationales at \emph{test} time, which would not be available in a real-world scenario. Still, we believe that sizable improvements would still be possible when using the gold rationales for training only.

\noindent Finally, we built a simple fact-checker, where the factuality of a claim is determined based on aggregating the predicted stances (using FNC's baseline system) of the documents we retrieved for it.
This yielded an accuracy 
of 56.2 when using the full documents, and 59.7 when using the best sentence + rationale (majority baseline of 50.5), thus confirming once again the utility of having a rationale, this time for a downstream task.

\section{Conclusion and Future Work}\label{sec:conclusion}

We have described a novel corpus that unifies stance detection, stance rationale, relevant document retrieval, and fact checking.
This is the first corpus to offer such a combination, not only for Arabic but in general.
We further demonstrated experimentally that these unified annotations, and the gold rationales in particular, are beneficial both for stance detection and for fact checking. 

In future work, we plan to cover other important aspects of fact checking such as source reliability, language style, and temporal information, which have been shown useful in previous research~\cite{Castillo:2011:ICT:1963405.1963500,lukasik-cohn-bontcheva:2015:ACL-IJCNLP,ma2016detecting,mukherjee2015leveraging,Popat:2017:TLE:3041021.3055133}.

\section*{Acknowledgment}
This research was carried out in collaboration between the MIT Computer Science and Artificial Intelligence Laboratory (CSAIL) and the HBKU Qatar Computing Research Institute (QCRI).

\bibliographystyle{acl_natbib}
\bibliography{references}

\end{document}